\documentclass[]{bytedance_seed}

\usepackage[toc,page,header]{appendix}
\usepackage{minitoc}
\usepackage{amsmath,amssymb}
\usepackage{wrapfig}
\usepackage{bbm}
\usepackage{xcolor}

\title{Scaling Reasoning Tokens via RL and Parallel Thinking:\\Evidence From Competitive Programming}

\author[2,\dagger]{Qianfan Zhang}
\author[3,\dagger]{Tianyu Guo}
\author[3,\dagger]{Xuandi Ren}
\author[4,\dagger]{Jiale Chen}
\author[1]{Ming Ding}
\author[1,\ddagger,*]{\\Ran Xin}
\author[1,*]{Xia Xiao}

\affiliation[1]{ByteDance Seed}
\affiliation[2]{Princeton University}
\affiliation[3]{UC Berkeley}
\affiliation[4]{Stanford University}

\contribution[\dagger]{Work done during internship at ByteDance Seed}
\contribution[~\ddagger]{Work done at ByteDance Seed}
\contribution[~*]{Supervising authors}

\abstract{
We study how to scale reasoning token budgets for competitive programming through two complementary approaches: training-time reinforcement learning (RL) and test-time parallel thinking. 
During RL training, we observe an approximately log-linear relationship between validation accuracy and the average number of generated reasoning tokens over successive checkpoints, and show two ways to shift this training trajectory: verification RL warmup raises the starting point, while randomized clipping produces a steeper trend in the observed regime.
As scaling single-generation reasoning during RL quickly becomes expensive under full attention, we introduce a multi-round parallel thinking pipeline that distributes the token budget across threads and rounds of generation, verification, and refinement. We train the model end-to-end on this pipeline to match the training objective to the test-time structure. Starting from Seed-OSS-36B, the full system with $16$ threads and $16$ rounds per thread matches the underlying RL model's oracle pass@16 at pass@1 using $7.6$ million tokens per problem on average, and surpasses GPT-5-high on 456 hard competitive programming problems from AetherCode.
}

\date{\today}
\correspondence{x.xiaxiao@bytedance.com}

\begin{document}
\maketitle

\vspace{-0.5cm}
\section{Introduction}

Scaling compute has been a central driver of progress in large language models and general AI systems~\citep{bitterlesson}. Pre-training scaling laws are well established \citep{kaplan2020scaling, hoffmann2022training}, and recent work has shifted toward scaling reasoning capabilities. On the training side, reinforcement learning (RL) has proven effective at incentivizing models to produce longer and more useful chain-of-thought reasoning traces \citep{wei2022chain}, as seen in systems such as OpenAI o1 \citep{o1} and DeepSeek-R1~\citep{deepseekr1}. On the test-time side, agentic pipelines that orchestrate multiple generations, verification, and search offer a complementary way to spend additional reasoning compute without increasing single-generation length \citep{snell2024scaling, geminideepthink}.

In this work, we study reasoning tokens scaling for competitive programming, a technical domain that remains challenging even for frontier models and provides unambiguous correctness signals through execution-based evaluation, making it an ideal testbed. Our contributions are the following:

\begin{itemize}
    \item We identify an \textbf{empirical log-linear trend} between average generated reasoning tokens and validation accuracy during RL training, and use it as a descriptive lens to compare RL variants. In this view, verification RL warmup raises the starting point, while randomized clipping yields a steeper trend.
    \item We introduce a \textbf{parallel thinking} framework, a multi-turn test-time pipeline that scales reasoning tokens across turns rather than within a single generation, combining multi-thread generation, self-verification, sequential self-refinement, and verification-based ranking. We train the model end-to-end on the full multi-turn pipeline via RL, aligning the training objective with the test-time structure.
    \item Starting from Seed-OSS-36B \citep{seedoss}, our full parallel thinking pipeline with $16$ threads and $16$ self-verify-refine rounds achieves \textbf{pass@1 accuracy matching the oracle pass@16} of the underlying RL model using \textbf{7.6 million tokens per problem on average}, and \textbf{surpasses GPT-5-high\footnote{GPT-5-high refers to GPT-5 evaluated with \texttt{reasoning\_effort=high} as described in the OpenAI GPT-5 System Card~\cite{gpt5}.}} on 456 hard competitive programming problems from AetherCode \citep{aethercode}.
\end{itemize}

\section{Scaling Reasoning Tokens via RL at Training Time}
\label{sec:rl}

We begin by studying how reasoning tokens scale during RL training. After describing the baseline setup in \Cref{sec:baseline}, we present an empirical log-linear trend between reasoning tokens and accuracy in \Cref{sec:loglinear}. This trend serves as a descriptive framework for comparing RL strategies. We then demonstrate in \Cref{sec:improving} two examples of improving the curve via verification RL warmup and randomized clipping. Finally, \Cref{sec:compute_wall} discusses the training-time compute wall that motivates our test-time approach.

\subsection{Baseline Setup}
\label{sec:baseline}

We use Seed-OSS-36B-Base \citep{seedoss} as the base model and train on proprietary competitive programming problems collected from online platforms like Codeforces.
For evaluation, we use AetherCode~\cite{aethercode} as the validation set, which contains 456 competitive programming problems collected from premier programming competitions such as IOI and ICPC.
Each solution is evaluated by execution against unit tests, receiving a reward of~$+1$ if it passes all tests and $0$ for any failure, including compilation error, wrong answer, and time limit exceeded. Responses that exceed the maximum context length are truncated and treated as incorrect.

Training uses asynchronous GRPO \citep{shao2024deepseekmath}, which is built on an in-house infrastructure similar to open-source RL frameworks \citep{verl, openrlhf, slime}, and allows $1$-step off-policy samples for improved training throughput. At each training step, the model generates $32$ rollouts for a batch of $16$ problems, receives rewards based on execution results, and updates the policy. Most RL runs use $256$ or $512$ A100 GPUs with maximum context length of $90$K tokens. Before RL training, we perform supervised fine-tuning (SFT) on approximately $6$K proprietary trajectories as a cold start. Total RL training data consists of approximately $10$K problems. 

\subsection{An Empirical Log-Linear Trend During RL Training}
\label{sec:loglinear}

\begin{wrapfigure}{r}{0.45\textwidth}
    \centering
    \vspace{-10pt}
    \hspace{-10pt}\includegraphics[width=0.42\textwidth]{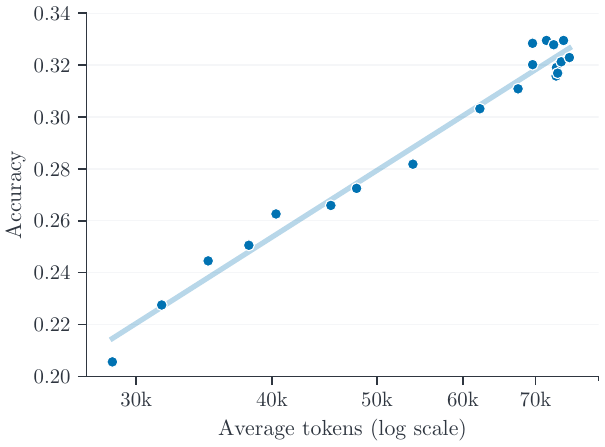}
    \vspace{-5pt}
    \caption{Log-linear trend: validation accuracy scales linearly with the logarithm of the average number of reasoning tokens during RL training. Each point corresponds to a successive RL training checkpoint.}
    \label{fig:loglinear}
    \vspace{-10pt}
\end{wrapfigure}

During RL training, the model progressively generates more reasoning tokens, consistent with prior observations in the literature~\citep{deepseekr1, prorl}. Across successive RL checkpoints, we observe that the relationship between the average generation length and validation accuracy is \textbf{log-linear}: accuracy increases linearly with the logarithm of the average token count, as shown in \Cref{fig:loglinear}.

We view this as an empirical regularity of our training setup rather than a universal law, and the plot should be interpreted as a compact summary of training dynamics in this regime. Still, the linear fit holds consistently across our different RL configurations and is stable enough to serve as a useful descriptive framework with several practical implications.
First, it enables early comparison of RL recipes, i.e., rather than training to convergence, one can fit the log-linear scaling curve from early checkpoints and extrapolate to compare the starting point and slope across different setups. Second, it provides a framework for diagnosing whether a new technique improves the intercept, the slope, or both, as we demonstrate in \Cref{sec:improving}. Third, it can guide compute budgeting by predicting the accuracy gain from a target increase in reasoning tokens.

\subsection{Improving the Log-Linear Trend}
\label{sec:improving}

We now illustrate two ways to improve the log-linear scaling curve. Randomized clipping steepens the slope by smoothing the hard reward boundary. Verification RL warmup raises the starting point before the generation RL stage described in \Cref{sec:baseline}. Other factors, such as the base model and the composition of the cold-start data, may also matter, but we do not explore them here.

\begin{figure}[h]
    \centering
    \includegraphics[width=1.0\textwidth]{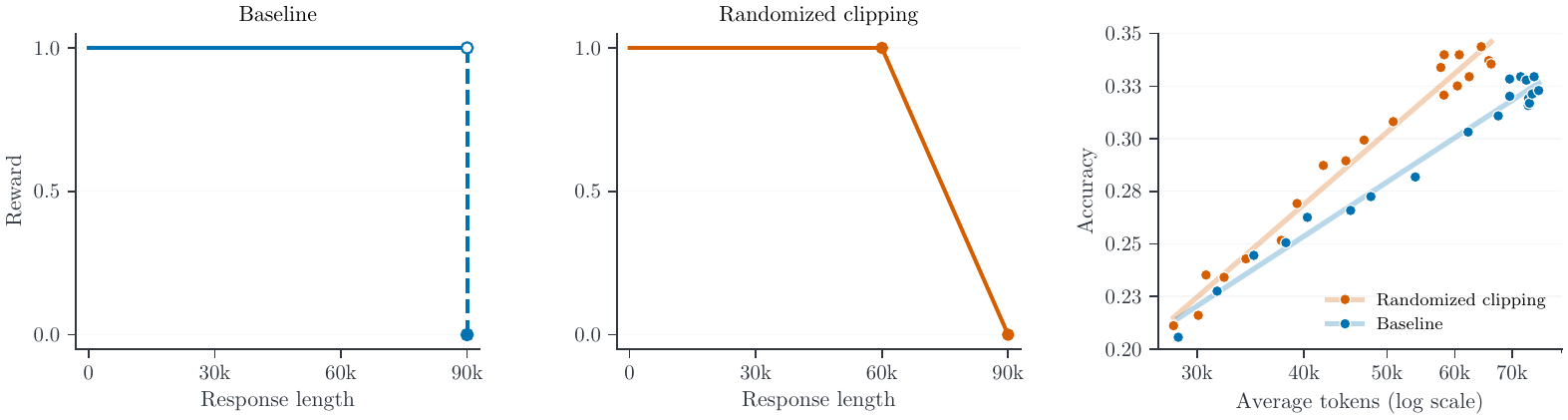}
    \caption{Randomized clipping replaces the hard reward cliff (left) with a smooth ramp (middle), producing a steeper log-linear scaling curve (right).}
    \label{fig:rc}
\end{figure}

\textbf{Steepening the slope via randomized clipping.} In the baseline generation RL setup from \Cref{sec:baseline}, each response is subject to a hard maximum token limit $L$. The reward for a prompt-response pair $(x, y)$ is
\begin{equation}
R^{(L)}(x,y) := \mathrm{score}(x,y) \cdot \mathbbm{1}[|y| \leq L],
\label{eq:hard_clip}
\end{equation}
where $\mathrm{score}(x,y) \in \{0, 1\}$ is the execution result and $|y|$ denotes the response length in tokens. For correct solutions, this is a step function in $|y|$ as shown in \Cref{fig:rc} (left), i.e., a response just below the limit receives full reward while one just above it receives none. The result is a sharp reward cliff near the boundary, with no gradual incentive to shorten near-limit responses.

We propose randomized clipping (RC), which replaces the fixed cap with a random cap sampled from a distribution $\mathcal{D}$. In expectation, this turns the hard cliff into a smooth penalty as a function of response length. The effective reward becomes
\begin{equation}
R^{(\mathcal{D})}(x,y) := \mathbb{E}_{L \sim \mathcal{D}}[R^{(L)}(x,y)] = \mathrm{score}(x,y) \cdot \mathbb{P}_{L \sim \mathcal{D}}(|y| \leq L) = \mathrm{score}(x,y) \cdot (1 - F_{\mathcal{D}}(|y|)),
\label{eq:rc}
\end{equation}
where $F_{\mathcal{D}}$ is the cumulative distribution function of $\mathcal{D}$. Because the penalty term $1 - F_{\mathcal{D}}(|y|)$ multiplies $\mathrm{score}(x,y)$, it only affects correct solutions; incorrect solutions still receive zero reward regardless of length. This preserves the model's incentive to explore longer reasoning on difficult problems. Different choices of $\mathcal{D}$ induce different smooth penalties: a Gaussian yields a sigmoid-shaped decay, a truncated exponential yields an exponential decay, and a uniform distribution yields a linear ramp. For simplicity, we use $\mathcal{D} = \mathrm{Uniform}(a, b)$, which gives the following piecewise linear reward:
\begin{equation}
R^{(\mathcal{D})}(x,y) = \mathrm{score}(x,y) \cdot \begin{cases} 1 & \text{if } |y| \leq a, \\[6pt] \dfrac{b - |y|}{b - a} & \text{if } a < |y| < b, \\[6pt] 0 & \text{if } |y| \geq b, \end{cases}
\label{eq:rc_uniform}
\end{equation}
We set $b = 90000$ to match the original hard limit and $a = 60000$ as a more economical budget within which most correct solutions already fit; see \Cref{fig:rc} (middle). As shown in \Cref{fig:rc} (right), RC steepens the log-linear scaling curve relative to the baseline, yielding better accuracy at a fixed reasoning token budget.

\textit{Remark.} While prior work has explored explicit length penalties added to the reward \citep{dapo, l1, laser}, RC offers a principled perspective rooted in randomized smoothing techniques from zeroth order optimization \citep{nesterov2017random, duchi2012randomized}: the effective penalty arises implicitly from randomizing the existing token limit, with the penalty shape controlled entirely by the choice of distribution $\mathcal{D}$.

\begin{figure}[h]
    \centering
    \vspace{-10pt}
    \includegraphics[width=0.8\textwidth]{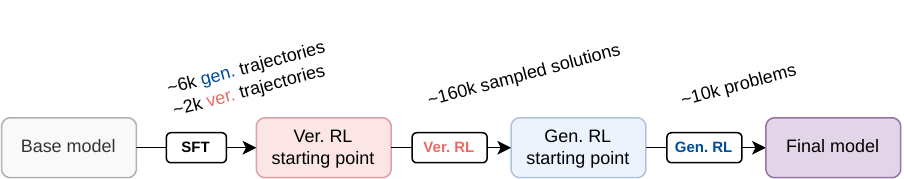}
    \caption{Training pipeline: SFT cold start with generation and verification trajectories, followed by verification RL and generation RL.}
    \label{fig:ver_pipeline}
\end{figure}

\textbf{Shifting the starting point via verification RL warmup.} Before the generation RL described in \Cref{sec:baseline}, we first train the model on a verification task using RL: given a problem and a candidate solution, predict whether the solution is correct. Here, the model produces a chain-of-thought analysis that traces execution, tests edge cases, and constructs counterexamples before outputting a binary verdict, where only the final verdict is used for scoring. After RL training, verification takes on average approximately 10K tokens, reflecting the depth of reasoning required to catch subtle algorithmic errors such as off-by-one flaws in combinatorial formulas or data structure misuse. \Cref{fig:ver_pipeline} illustrates the full training pipeline.

In particular, the SFT cold start of verification RL uses approximately 6K generation trajectories alongside 2K verification trajectories created by rejection sampling from a generation RL checkpoint. Verification RL then trains on approximately 160K candidate solutions collected from previous generation RL runs, each automatically labeled correct or incorrect by execution against unit tests. The reward is $+1$ for a correct verdict and $0$ otherwise, using the same RL configuration as generation training described in \Cref{sec:baseline}.

\begin{figure}[h]
    \centering
    \includegraphics[width=0.4\textwidth]{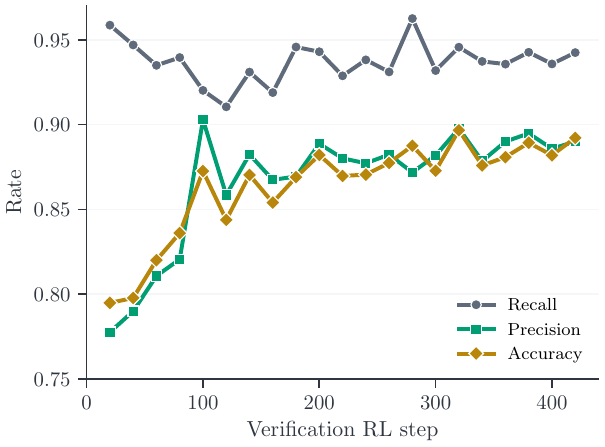}
    \qquad\qquad
    \includegraphics[width=0.4\textwidth]{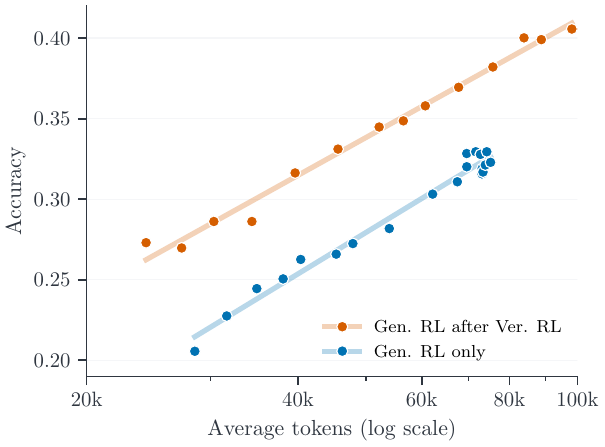}
    \caption{Verification RL warmup. \textbf{Left}:~recall remains high while precision and accuracy improve during verification RL. \textbf{Right}:~initializing generation RL from the verification checkpoint shifts the log-linear scaling curve upward.}
    \label{fig:ver_warmup}
\end{figure}

The verification RL training dynamics is shown in \Cref{fig:ver_warmup} (left). Recall starts high at $\sim$0.96 and remains stable throughout training, while precision climbs steadily from $\sim$0.78 to $\sim$0.89. This indicates that the SFT checkpoint already detects most correct solutions, but frequently misclassifies incorrect ones as correct. Verification RL primarily reduces these false positives. By step 420, accuracy reaches $\sim$0.89.

The resulting verification RL checkpoint then serves as initialization for generation RL. As shown in \Cref{fig:ver_warmup} (right), verification RL warmup shifts the log-linear scaling curve upward: generation RL initialized from the verification checkpoint achieves higher accuracy than generation RL alone at any given token budget. At the standard $90$K maximum context length, generation RL without verification warmup plateaus at ${\sim}0.33$ once the average generation length reaches ${\sim}70$K tokens, whereas the warm-started run reaches ${\sim}0.38$ using a similar number of tokens. Continuing the warm-started run with an extended $120$K maximum context length allows it to keep improving beyond the original budget range while remaining consistent with the same overall trend. We hypothesize that verification training improves the model's ability to internally evaluate solution correctness, yielding a stronger starting point and thus a more favorable scaling trajectory for subsequent generation RL.

\textit{Remark.} The success of verification RL warmup suggests a broader principle: incorporating data that exercises sub-capabilities useful for generation may further improve the generation scaling curve. We leave systematic exploration of this direction to future work.

\subsection{The Compute Wall}
\label{sec:compute_wall}

The empirical log-linear trend suggests that, within the range we observe, longer generations are associated with higher validation accuracy. However, directly pushing sequence length further during RL training quickly becomes impractical.
With full attention, the computational cost scales quadratically with sequence length. At an average response length of $\sim$100K tokens, a single RL training step takes approximately 4 hours on 256 A100s, making further scaling prohibitively expensive. Efficient attention mechanisms \cite{linearattention, sparseattention} might alleviate this bottleneck, but this compute wall motivates our test-time approach described in the next section.

\section{Scaling Reasoning Tokens via Parallel Thinking at Test Time}
\label{sec:parallel}

To scale reasoning tokens beyond the compute wall identified in \Cref{sec:compute_wall}, we introduce a parallel thinking framework that distributes the reasoning budget across independent threads, each executing multiple rounds of self-verification and self-refinement. Each individual generation remains short, sidestepping the quadratic attention bottleneck, while the total generated tokens extend by over an order of magnitude beyond what is feasible at training time. We describe the inference pipeline in \Cref{sec:pipeline_overview}, the training procedure that aligns the model with this multi-round structure in \Cref{sec:e2e_rl}, and the resulting scaling behavior in \Cref{sec:scaling_results}.

\subsection{Pipeline Overview}
\label{sec:pipeline_overview}

\begin{figure}[h]
    \centering
    \vspace{-10pt}
    \includegraphics[width=0.7\textwidth]{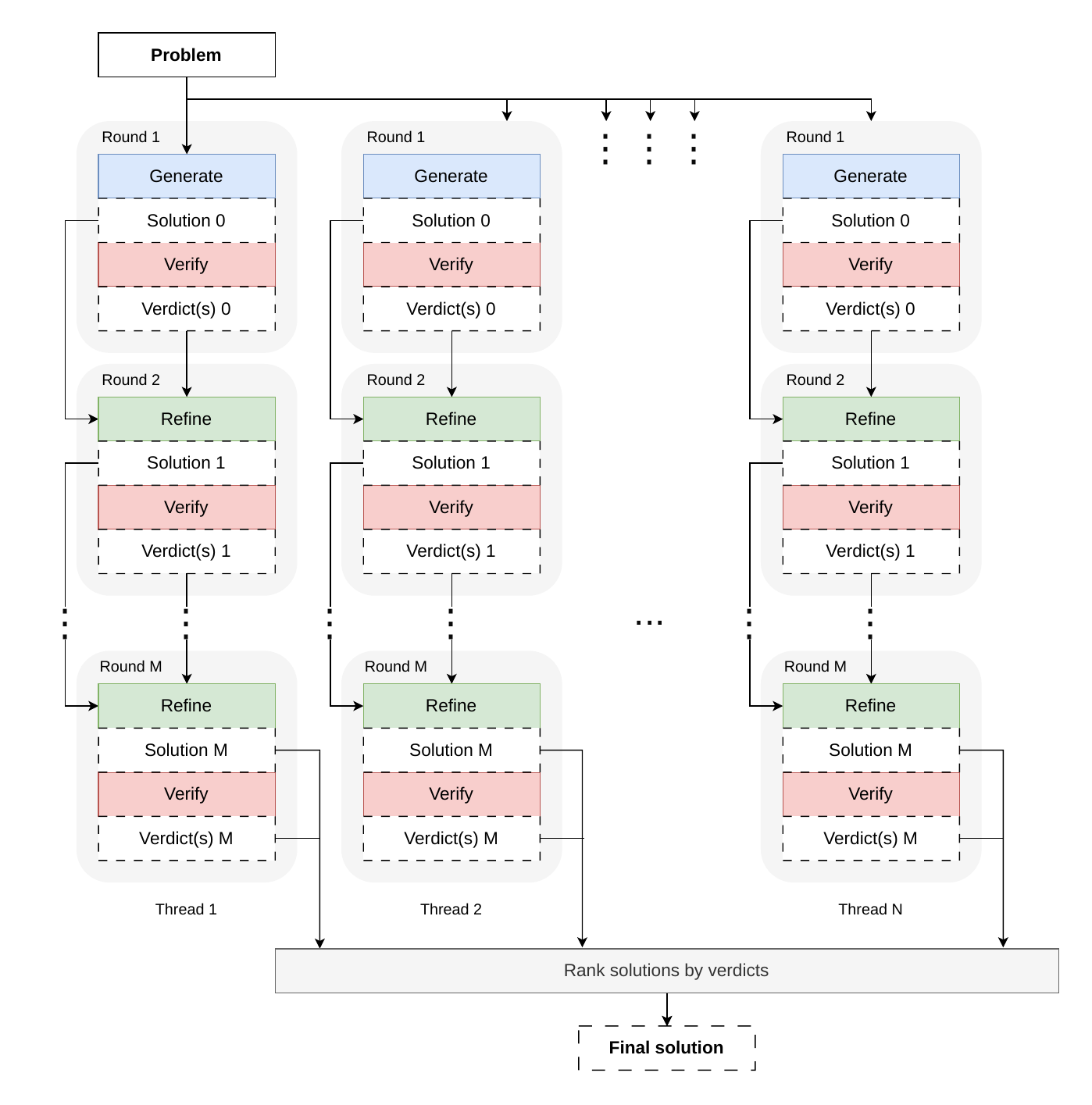}
    \vspace{-10pt}
    \caption{The parallel thinking inference pipeline. The system spawns $N$ independent threads. Each thread \textbf{generates} a candidate solution, then \textbf{verifies} it via $V$ independent sampling calls, each producing a correctness judgment and reasoning. If all $V$ verdicts unanimously deem the solution correct, the thread terminates early; otherwise, one negative verdict is randomly selected and the model \textbf{refines} the solution conditioned on the previous attempt and the selected reasoning. This verify-refine loop repeats for up to $M$ rounds. After all threads complete, all solutions across threads and rounds are ranked by verification vote count, and the highest-scoring one is returned.}
    \label{fig:pipeline}
\end{figure}

We now detail the test-time inference procedure; see \Cref{fig:pipeline} for an overview. Given a problem $x$, the parallel thinking system spawns $N$ independent threads. Each thread executes up to $M$ rounds, where each round consists of the following two steps:
\begin{enumerate}
    \item \textbf{Generate/Refine}: In the first round, the model produces a candidate solution from scratch. In subsequent rounds, the model generates a refined solution conditioned on the previous attempt and a negative verification verdict from the previous round.
    \item \textbf{Verify}: Given a solution $y$, the model produces $V$ verdicts via $V$ independent sampling calls, each containing a correctness judgment and its corresponding reasoning. If all $V$ verdicts unanimously deem the solution correct, the thread terminates early.
\end{enumerate}
The total reasoning token budget is at most $N \times M \times T_{\mathrm{round}}$, where $T_{\mathrm{round}}$ is the average tokens per round, and is often less due to early termination. After all threads complete, the system selects a final answer from all solutions produced across all threads and rounds. Let $y_{n,m}$ denote the solution produced at round $m$ of thread $n$, with verification verdicts $v_{n,m,j} \in \{0, 1\}$ for $j = 1, \ldots, V$. Each solution is scored by its number of positive verdicts:
\begin{equation}
    s_{n,m} = \sum_{j=1}^{V} v_{n,m,j}.
    \label{eq:score}
\end{equation}
The solution with the highest score is selected as the final answer, with ties broken by preferring earlier rounds and remaining ties broken randomly. We prefer earlier rounds because refinement can degrade correct solutions when guided by incorrect verdicts. Among equally scored solutions, the one requiring fewer refinement steps is more likely to be correct.

\textbf{Why self-verification?} Unlike mathematical reasoning where majority voting on final answers is effective~\citep{wang2023selfconsistency}, competitive programming solutions are code and cannot be meaningfully compared by output equivalence alone. An alternative is to check candidates against brute-force solvers with self-generated unit tests~\citep{alphacode}, but producing correct-by-construction test generators and reference solutions is itself a challenging problem. We instead rely on self-verification, where the same model trained via verification RL, as shown in \Cref{sec:improving}, judges each candidate, providing both the ranking signal for selecting among candidate solutions and the feedback that drives refinement, with no external oracle required.

\textbf{Beyond independent threads.} Our pipeline can be naturally extended in several directions: (1)~incorporating unit-test-based verification oracle alongside self-verification, as our preliminary experiments show this further improves performance; (2)~using unit test execution signals, such as failing test cases, to directly guide solution refinement; (3)~replacing the linear verify-refine chain with tree search over refinement paths, using verification scores to prune and expand branches~\citep{restmcts, bfsprover1}; (4)~crossing solutions and verification reasoning across threads via summarization, akin to a neural evolutionary algorithm~\citep{elm, funsearch, alphaevolve}; and (5)~combining the vote-count ranking with more fine-grained selection strategies such as pairwise comparison or learned ranking models. In this work, we focus on the simplest instantiation, i.e., independent threads with linear verify-refine chains, and show that it already yields strong scaling. We leave systematic exploration of the aforementioned richer strategies to future work.

\subsection{Training for Parallel Thinking}
\label{sec:e2e_rl}

\textbf{Separate training.} The simplest approach is rather to deploy the checkpoint produced by the verification~RL followed by generation RL pipeline in \Cref{sec:improving}. However, its training context is misaligned with the multi-round test-time pipeline in \Cref{sec:pipeline_overview}. In particular, each capability is trained independently on single-round contexts, whereas at inference the verifier must evaluate on-policy solutions from the current generator, and refinement must be conditioned on the verification critique of the current solution. This mismatch between training and inference contexts motivates end-to-end RL training on the full multi-round~pipeline.

\textbf{End-to-end RL.} To close this train--test gap, we train the model end-to-end on the full multi-round pipeline via RL, starting from the verification RL checkpoint of \Cref{sec:improving}. For each problem $x$, we sample a group of trajectories, each consisting of $M$ rounds of alternating turns:
\begin{equation}
    \tau = (\underbrace{y_{\texttt{gen}}^1,\, y_{\texttt{ver}}^1}_{\text{round } 1},\, \underbrace{y_{\texttt{ref}}^2,\, y_{\texttt{ver}}^2}_{\text{round } 2},\, \ldots,\, \underbrace{y_{\texttt{ref}}^M,\, y_{\texttt{ver}}^M}_{\text{round } M}),
    \label{eq:trajectory}
\end{equation}
where $y_{\texttt{role}}^i$ denotes the model output at round $i$ with the given role. Each round consists of a generation or refinement turn followed by a verification turn. Unlike test time where each verification turn produces $V$ independent verdicts, training uses a single verdict ($V=1$) to reduce rollout cost. Early termination follows the same rule, i.e., refinement is skipped if the verdict deems the solution correct. We now describe the reward structure and policy optimization.

\textbf{Per-turn rewards.} Each turn~$t$ is optimized for its own immediate reward $r_t$, determined by its role:
\begin{equation}
    r_t = \begin{cases}
        \mathrm{exec}(x, y_t) & \text{if } y_t \text{ is a generation or refinement step}, \\[4pt]
        \mathbbm{1}\!\big[\mathrm{verdict}(y_t) = \mathrm{exec}(x, y_{t-1})\big] & \text{if } y_t \text{ is a verification step},
    \end{cases}
    \label{eq:per_turn_reward}
\end{equation}
where $\mathrm{exec}(x, y) \in \{0, 1\}$ is the execution result of solution $y$ against the test suite for problem $x$, $\mathrm{verdict}(y_t) \in \{0, 1\}$ is the binary correctness prediction extracted from the verification output~$y_t$, and $y_{t-1}$ is the solution being verified. In other words, generation and refinement are rewarded for producing correct code, while verification is rewarded for predicting correctness accurately. Training on multi-round rollouts aligns the context distribution with the test-time pipeline, i.e., each turn is optimized in the presence of all preceding turns, so the model learns to generate, verify, and refine in context.

\textit{Remark.} Optimizing the policy based on the immediate reward of each turn makes the underlying problem a multi-task contextual bandit~\citep{deshmukh2017multitask}. A natural extension is to propagate credit across turns via a discounted return $R_t = \sum_{k=t}^{T} \gamma^{k-t} r_k$ with $\gamma > 0$, so that the return of each turn reflects downstream outcomes. For instance, a generation step would be rewarded not only for immediate correctness but also for producing solutions that are easy to diagnose via verification when incorrect, and the verifier would receive gradients that reflect how its feedback is used during refinement. Here, we adopt $\gamma = 0$ for simplicity, because propagating credit across turns introduces additional hyperparameter sensitivity. We leave that direction to future work.

\textbf{Policy optimization.} Since generation, verification, and refinement have heterogeneous reward distributions, advantage normalization is non-trivial. We consider the following two approaches.

\begin{itemize}
    \item \textbf{REINFORCE with batch-level whitening.} The advantage of turn $t$ is $A_t = (r_t - \mu)/(\sigma + \delta)$, where $\mu$ and~$\sigma$ are the mean and standard deviation of all rewards $\{r_k\}$ in the batch regardless of role. This treats all turns uniformly, but normalizing across roles with different reward distributions may distort their respective advantage and gradient signals.
    \item \textbf{Turn-grouped GRPO.} Standard GRPO~\citep{shao2024deepseekmath} is a contextual bandit algorithm that computes advantages within a group of responses to the same prompt. Here, we extend it by computing advantages separately for each turn. The advantage of turn $t$ in a given trajectory is
    \begin{equation}
        A_t = r_t - \mu_t, \quad \mu_t = \mathbb{E}_{\text{batch}}[r_t],
        \label{eq:advantage}
    \end{equation}
    where $\mu_t$ is the mean reward at turn $t$ across all trajectories in the batch. Note that we do not divide by the turn-grouped standard deviation in~\Cref{eq:advantage}, as mean-centering alone suffices.
    Since each turn $t$ maps to a fixed role, this calibrates the gradient signal per turn and role independently, avoiding the distortion from batch-level whitening.
\end{itemize}

In both cases, the policy is updated using the clipped surrogate objective~\citep{schulman2017ppo}:
\begin{equation}
    \mathcal{L}(\theta) = -\,\mathbb{E}_{t}\Big[\min\!\big(\rho_t(\theta)\, A_t,\;\; \mathrm{clip}(\rho_t(\theta), 1-\varepsilon, 1+\varepsilon)\, A_t\big)\Big],
    \label{eq:ppo_objective}
\end{equation}
where $\rho_t(\theta) = \pi_\theta(y_t \mid c_t) / \pi_{\theta_{\text{old}}}(y_t \mid c_t)$ is the importance sampling ratio between the current and previous policies, $\varepsilon$ is the clipping parameter, and $c_t$ denotes the context at turn $t$, comprising the problem statement and respective model outputs from preceding turns.

\subsection{Scaling Results}
\label{sec:scaling_results}

\begin{figure}[h]
    \centering
    \includegraphics[width=0.47\textwidth]{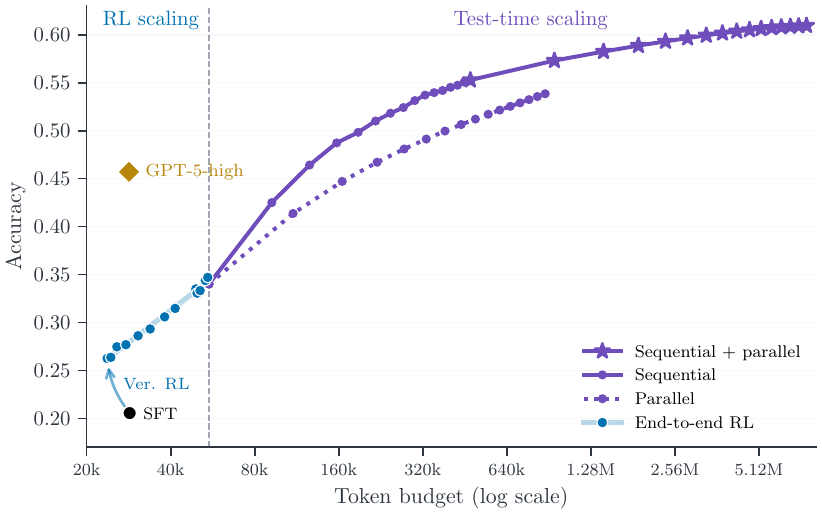}
    \qquad
    \includegraphics[width=0.47\textwidth]{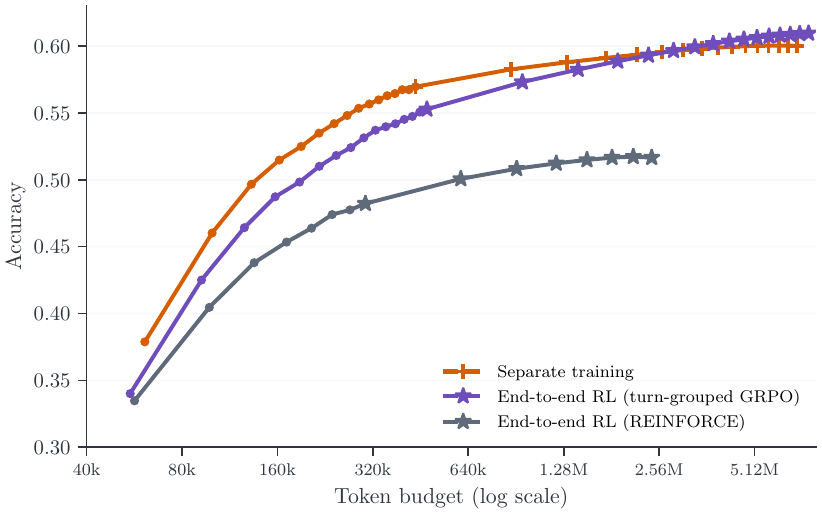}
    \caption{\textbf{Left}: Unified token budget axis spanning \textcolor[HTML]{0072B2}{training-time RL scaling} from 20K to 60K tokens and \textcolor[HTML]{6F4EBC}{test-time parallel thinking} with the end-to-end RL model from 60K to 7.6M tokens. Sequential refinement outperforms parallel generation at every token budget; combining both extends scaling beyond the sequential plateau and surpasses GPT-5-high by a wide margin. \textbf{Right}: Effect of end-to-end RL on sequential-plus-parallel scaling. \textcolor[HTML]{D55E00}{Separate training} initially outperforms \textcolor[HTML]{6F4EBC}{end-to-end RL with turn-grouped GRPO} due to stronger single-round generation, but the latter catches up by ${\sim}2$M tokens and overtakes it at higher budgets. \textcolor[HTML]{5F6B7A}{End-to-end RL with REINFORCE} scales substantially~worse.\protect\footnotemark}
    \label{fig:scaling_results}
\end{figure}

We now evaluate how parallel thinking scales with reasoning tokens. All experiments use the end-to-end RL model trained with turn-grouped GRPO from \Cref{sec:e2e_rl} with $V=8$ verdicts per verification step unless noted otherwise. The starting point for all test-time scaling curves is a single generation from the end-to-end RL checkpoint, at ${\sim}0.34$ accuracy and ${\sim}60$K tokens. \Cref{fig:scaling_results}~(left) places RL training-time scaling and test-time parallel thinking on a unified token budget axis, and \Cref{fig:scaling_results}~(right) isolates the effect of end-to-end RL. Additional ablations on the effect of varying the number of threads and verdicts are provided in \Cref{app:scaling_ablations}. A few observations are in~order.

\footnotetext{Due to limited computational resources, we evaluate the REINFORCE checkpoint only up to $N{=}8$ threads and $M{=}8$ rounds. Even within this truncated range, it remains substantially below the other methods at comparable token budgets.}

\textbf{Sequential refinement is more token-efficient than parallel generation.} Sequential scaling fixes the number of threads to $N{=}1$ and increases the number of verify-refine rounds from $M{=}1$ to $M{=}16$. This reaches ${\sim}0.49$ by $160$K tokens, surpassing GPT-5-high~\cite{gpt5} at ${\sim}0.46$ within a few refinement rounds, and continues to climb to ${\sim}0.55$ at ${\sim}500$K tokens before plateauing. This steep trajectory reflects the core advantage of sequential scaling, i.e., each refinement round is conditioned on verification feedback from the previous attempt, so the model can diagnose and correct specific errors adaptively rather than regenerating blindly.

\textbf{Parallel scaling is less token-efficient but reduces wall-clock time.} Parallel scaling fixes the number of rounds to $M{=}1$ and increases the number of threads from $N{=}1$ to $N{=}16$, with each thread generating a single solution that is then ranked by verification score. This consistently lags sequential refinement at the same token budget, e.g., ${\sim}0.45$ versus ${\sim}0.49$ at $160$K tokens. This gap is expected because parallel threads generate independently without leveraging verification feedback, so they behave like best-of-$N$ with a learned ranker. The practical value of parallelism lies in wall-clock latency reduction rather than token efficiency, since all threads execute concurrently.

\textbf{Combining sequential and parallel scaling yields the best results.} Sequential-plus-parallel scaling first applies $M{=}16$ verify-refine rounds within each thread, then increases the number of threads from $N{=}1$ to $N{=}16$. At the full configuration of $N{=}16$ threads and $M{=}16$ rounds, accuracy reaches ${\sim}0.61$ at $7.6$M tokens, well beyond the sequential-only plateau of ${\sim}0.55$. This demonstrates that parallelism adds complementary value once sequential refinement within each thread has saturated, i.e., different threads explore different solution strategies, and verification-based ranking selects the best one across all threads and rounds. The full pipeline surpasses GPT-5-high~\cite{gpt5} by a substantial margin.

\textbf{End-to-end RL overtakes separate training at scale.} \Cref{fig:scaling_results}~(right) compares the separately trained model from \Cref{sec:improving} with the end-to-end RL model under sequential-plus-parallel scaling. The separately trained model\footnote{We use an earlier checkpoint from the separately trained pipeline with ${\sim}0.38$ accuracy and ${\sim}65.5$K average generation length, rather than the final checkpoint which exceeds $100$K tokens and makes multi-round rollouts prohibitively slow. See \Cref{fig:ver_warmup} (right) for the corresponding scaling curve.} starts ${\sim}4$ points higher at $65.5$K tokens, $0.38$ versus $0.34$, reflecting its stronger single-generation quality from independently optimized generation and verification stages. However, the end-to-end RL model catches up by ${\sim}2$M tokens and reaches ${\sim}0.61$ at the highest budgets, slightly exceeding the separately trained model at ${\sim}0.60$. While the final gap is modest, the scaling trajectory is qualitatively different. In particular, end-to-end training trades single-round quality for improved multi-round coherence, as the verifier learns to critique on-policy generations and the refiner learns to act on on-policy verification feedback, a coordination that separate training cannot achieve. \Cref{fig:scaling_results}~(right) also shows that turn-grouped GRPO substantially outperforms REINFORCE with batch-level whitening at ${\sim}0.61$ versus ${\sim}0.51$ at the highest budgets, confirming that calibrating advantages per turn and role is important for multi-round training.

\section{Related Work}
\label{sec:related}

\textbf{RL for LLM reasoning.} Chain-of-thought (CoT) prompting \citep{wei2022chain} establishes that LLMs benefit from generating intermediate reasoning steps before producing a final answer. RL extends this paradigm by training models to produce much longer and more effective reasoning traces. OpenAI o1 \citep{o1} first demonstrates large-scale RL for reasoning, and later DeepSeek-R1 \citep{deepseekr1} shows that RL alone can incentivize long CoT directly from a base model without any cold-start. DeepSeekMath \citep{shao2024deepseekmath} introduces GRPO for mathematical reasoning, which we adapt as our core policy optimization algorithm. DeepSeekMath-V2 \citep{deepseekmath2} extends this line by training LLM verifiers for self-verifiable mathematical reasoning, relying on human annotations to supervise verification. Relevant work explores various RL recipes, including dynamic sampling, clip-higher, REINFORCE++, prolonged training schedules, entropy stabilization, and length control \citep{dapo, reinforcepp, prorl, entropyrl, l1}. Kimi k1.5 \citep{kimik15} employs a variant of online mirror descent for policy optimization and jointly trains on text and vision data for multimodal reasoning, while Qwen3~\citep{qwen3} introduces hybrid thinking that dynamically switches between thinking and non-thinking modes. The influence of the base model on long CoT RL training is also studied, with work questioning whether RL genuinely expands reasoning beyond base model capabilities \citep{leaprl}, investigating how different base models respond to zero-RL training \citep{simplerlzoo}, and analyzing the training dynamics of R1-Zero-like approaches \citep{r1zero}. 
Our work documents an empirical log-linear trend between accuracy and reasoning tokens during RL training on competitive programming, and introduces randomized clipping and verification RL warmup as two techniques that improve the observed trend.

\textbf{Test-time compute scaling.} Test-time compute can be scaled through several complementary mechanisms. Best-of-$N$ sampling and majority voting \citep{wang2023selfconsistency} generate multiple independent solutions, while tree search methods, including Monte Carlo tree search \citep{restmcts, alphaproof} and best-first search \citep{bfsprover1, bfsprover2}, explore branching paths. In competitive programming, AlphaCode \citep{alphacode} pioneers competition-level code generation by generating up to one million samples per problem and clustering them by behavioral equivalence on generated test inputs, while AlphaCode~2 \citep{alphacode2} achieves improved performance by leveraging a stronger base model and learned scoring. Process reward models provide step-level supervision to guide search, using human annotations \citep{lightman2023lets}, automated labeling \citep{mathshepherd}, or implicit rewards derived from outcome signals \citep{prime}. Recent work shows that optimally allocating test-time compute can be more effective than scaling model parameters \citep{snell2024scaling}, and meta RL can further improve token efficiency at test time \citep{qu2025optimizing}. SCoRe \citep{score} and Goedel-Prover-V2 \citep{goedelprover2} train self-correction via multi-turn RL. Seed-Prover \citep{seedprover} scales test-time compute by decomposing proofs into lemmas cached in a reusable memory pool. Our parallel thinking framework combines parallel generation with sequential self-verification and self-refinement, and trains the model end-to-end on the multi-turn pipeline via RL.
\section{Conclusion}
\label{sec:conclusion}

We studied scaling reasoning tokens for competitive programming through two complementary mechanisms.
At training time, we observe an approximately log-linear relationship between validation accuracy and average generated tokens over successive RL checkpoints, and we demonstrate two examples of improving this trend: randomized clipping steepens the slope, while verification RL warmup shifts the starting point upward by training atomic sub-capabilities of generation.
Since the quadratic attention cost imposes a compute wall on single-generation length, we introduced a parallel thinking framework that scales reasoning tokens across $N$ independent threads, each executing up to $M$ rounds of verify-refine, extending the effective token budget by over an order of magnitude. Sequential refinement outperforms parallel generation at every token budget, and their combination yields the best results. End-to-end co-training further improves performance by aligning the training objective with the test-time structure. Starting from Seed-OSS-36B-Base, the full pipeline achieves pass@1 accuracy matching the oracle pass@16 of the underlying RL model using $7.6$M tokens per problem on average, and surpasses GPT-5-high on 456 hard problems from AetherCode. By unifying training-time RL and test-time search along a common reasoning token budget axis, our framework provides a scalable path for improving reasoning on complex tasks.


\bibliographystyle{plainnat}
\bibliography{main}

\newpage
\beginappendix

\section{Ablation on Number of Verdicts in Parallel Thinking}
\label{app:scaling_ablations}

In this section, we ablate how the number of verification verdicts $V$ affects the accuracy of the parallel thinking pipeline described in \Cref{sec:parallel}. All experiments use an early checkpoint of the end-to-end RL model trained with turn-grouped GRPO and fix $M{=}1$, i.e., no refinement, isolating the contribution of verification from sequential scaling.

\begin{figure}[h]
    \centering
    \includegraphics[width=0.5\textwidth]{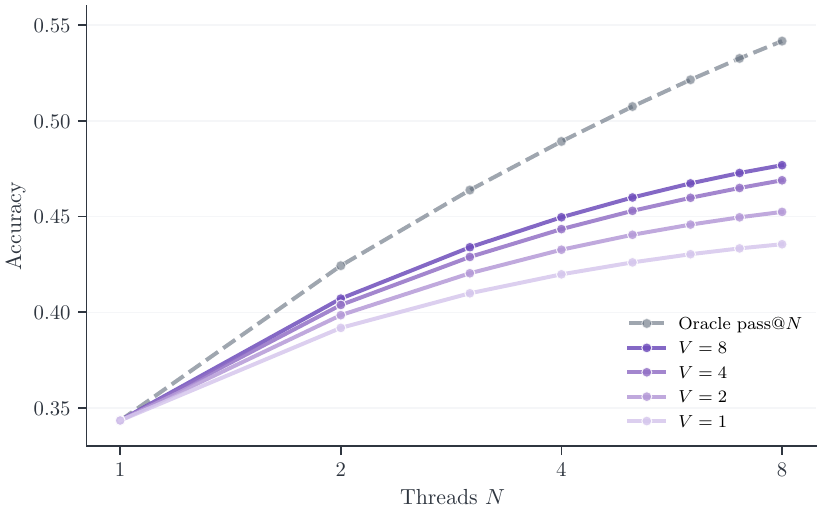}
    \caption{Parallel scaling accuracy versus number of threads $N$ for varying verdicts $V \in \{1, 2, 4, 8\}$ (\textcolor[HTML]{6F4EBC}{colored}),~with $M{=}1$, i.e., no refinement. More verdicts improve ranking quality, but the gap to the oracle pass@$N$ (\textcolor[HTML]{5F6B7A}{dashed}) remains substantial, suggesting that verifier capability is the primary bottleneck.}
    \label{fig:parallel_ablation}
\end{figure}

\Cref{fig:parallel_ablation} shows accuracy as a function of threads $N$ for $V \in \{1, 2, 4, 8\}$ verdicts. Increasing $V$ consistently improves accuracy at every thread count, but the gains are modest: at $N{=}8$, moving from $V{=}1$ to $V{=}8$ improves accuracy from ${\sim}0.43$ to ${\sim}0.48$, while the oracle pass@$8$ sits at ${\sim}0.54$. The persistent gap indicates that the verification capability to distinguish correct from incorrect solutions, rather than the number of ranking samples, is the primary bottleneck. Prolonged verification RL training and more diverse verification data, for instance, are promising directions for closing this gap. This observation also motivates the sequential refinement approach in \Cref{sec:scaling_results}, which uses the diagnostic information in negative verdicts to improve solutions rather than only rank them.

\newpage
\section{Oracle Pass Rate Scaling}
\label{app:oracle_pass_rate}

\begin{figure}[h]
    \centering
    \includegraphics[width=0.55\textwidth]{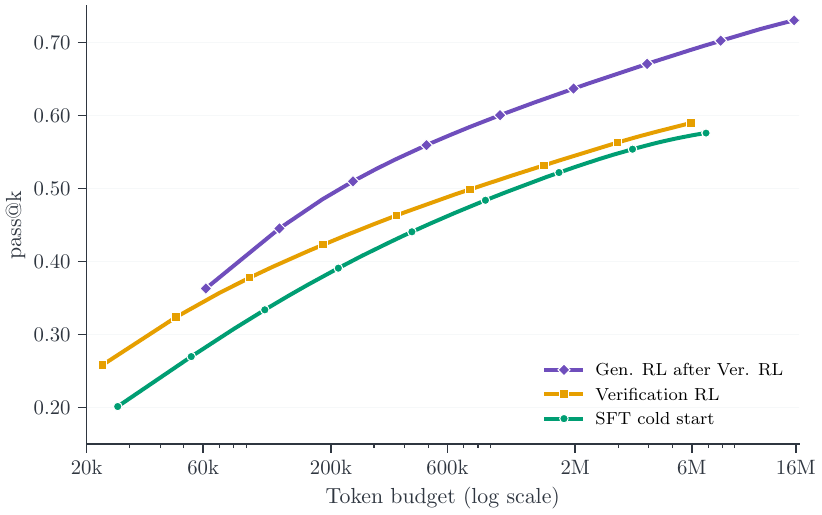}
    \caption{Oracle pass@$k$ versus token budget for three training stages: \textcolor[HTML]{009E73}{SFT cold start}, \textcolor[HTML]{E69F00}{verification RL}, and \textcolor[HTML]{6F4EBC}{generation RL after verification RL}. Generation RL dominates at every token budget. The gap persists even at pass@$256$, indicating that RL can expand solution coverage beyond the SFT frontier rather than merely concentrating probability on existing solutions.}
    \label{fig:oracle_passk}
\end{figure}

In this section, we evaluate the oracle pass@$k$ scaling behavior of three key checkpoints in our training pipeline: (1) the SFT cold-start checkpoint, (2) the verification RL checkpoint, and (3) the generation RL checkpoint initialized from verification RL\footnote{As in \Cref{sec:scaling_results}, we use an earlier generation RL checkpoint with ${\sim}0.38$ accuracy and ${\sim}65.5$K average generation length rather than the final one, which exceeds $100$K tokens. See \Cref{fig:ver_warmup} (right) for the corresponding scaling curve.}, i.e., the full pipeline of \Cref{fig:ver_pipeline}. We sample up to $k{=}256$ generations per problem, and the token budget is computed as $k$ times the average generation length. Experimental results are shown in \Cref{fig:oracle_passk} and two remarks are in order.

\textbf{Verification RL improves per-sample quality but not coverage.} The verification RL checkpoint, despite not being trained on generation, improves over SFT at low $k$, e.g., ${\sim}0.26$ versus ${\sim}0.20$ at pass@$1$, consistent with our hypothesis in \Cref{sec:improving} that verification training transfers evaluative capabilities that benefit generation. However, the two checkpoints converge to ${\sim}0.57$--$0.59$ at pass@$256$, suggesting that verification RL increases per-sample success rate rather than expanding the set of reachable solutions by, e.g., avoiding common errors.

\textbf{Generation RL expands solution coverage.} The generation RL checkpoint maintains a substantial lead at ${\sim}0.73$ pass@$256$. Since the gap persists even at large $k$ where sampling diversity is high, generation RL genuinely expands solution coverage beyond what SFT can reach at any sampling budget. This contrasts with the finding of~\cite{leaprl}, which observes that RL does not expand reasoning beyond the base model on mathematical reasoning tasks. A possible explanation is that competitive programming, with its combinatorially richer solution space and execution-based reward signal, provides a stronger training signal for RL to discover novel strategies outside the SFT distribution.

\newpage
\section{Prompt Templates}
\label{app:prompts}

We provide the prompt templates used for each role in the parallel thinking pipeline described in \Cref{sec:parallel}. In all templates, \texttt{\{problem\}} denotes the problem statement including input/output format and examples, \texttt{\{solution\}} denotes the candidate solution with its explanation and code, and \texttt{\{verdict\_reasoning\}} denotes the verification reasoning from the previous round.

\subsection*{Generation}

\begin{quote}
\small\ttfamily\raggedright
You are solving the given programming contest problem with a C++ solution.\par\medskip
\{problem\}
\end{quote}

\subsection*{Verification}

\begin{quote}
\small\ttfamily\raggedright
You are given a programming contest problem and a proposed solution. Your task is to determine whether the solution is correct (should receive Accepted) or incorrect (e.g., Wrong Answer, Time Limit Exceeded, Runtime Error, etc.).\par\medskip
Important requirements:\par
\hangindent=1em -- Carefully reason about all edge cases and constraints.\par
\hangindent=1em -- If you decide the solution is incorrect, you MUST identify at least one clear reason, such as a logical flaw, missing case, incorrect complexity, or a specific counterexample.\par
\hangindent=1em -- A counterexample should be described concretely (e.g., a specific input and what goes wrong).\par
\hangindent=1em -- Do NOT hedge: pick exactly one verdict, Correct or Incorrect.\par\medskip
Your response MUST follow EXACTLY this format (with no extra text before or after):\par
Line 1: "Verdict: Correct." or "Verdict: Incorrect."\par
Line 2+: One or few short paragraphs explaining the reasoning for that verdict. If Incorrect, you MUST mention at least one specific failing scenario, logical flaw, or counterexample.\par\medskip
\{problem\}\par\medskip
\{solution\}
\end{quote}

\subsection*{Refinement}

\begin{quote}
\small\ttfamily\raggedright
You are correcting a programming contest submission. A judge provided a verdict explaining why the prior attempt failed. Read the feedback carefully and emit an improved C++ solution. The judge feedback may be noisy and the previous solution might actually be correct, but you must still output a solution using the same format (solution explanation followed by reference code).\par\medskip
\{problem\}\par\medskip
\{solution\}\par\medskip
\{verdict\_reasoning\}
\end{quote}

\end{document}